\newif\ifisfinal\isfinaltrue
\documentclass[11pt,a4paper]{article}

\ifisfinal
\usepackage[hyperref]{emnlp2021}
\else
\usepackage[hyperref,review]{emnlp2021}
\fi
\usepackage{zslib}
\usepackage{zenlist}

\acsetup{
  first-long-format=\textnormal,
  hyperref=false,
  tooltip=false,
}

\title{\doclongtitle}
\author{\theauthor\ \and Barbara Di Eugenio \and Cornelia Caragea
  \AND
  \texttt{\{plande2, bdieugen, cornelia\}@uic.edu} \\
  \theorganizationdept\\
  \theorganization
}
\date{}

\begin{document}

\maketitle

\begin{abstract}
  Reproducing results in publications by distributing publicly available source
  code is becoming ever more popular.  Given the difficulty of reproducing \ml\
  experiments, there have been significant efforts in reducing the variance of
  these results.  As in any science, the ability to consistently reproduce
  results effectively strengthens the underlying hypothesis of the work, and
  thus, should be regarded as important as the novel aspect of the research
  itself.  The contribution of this work is a framework that is able to
  reproduce consistent results and provides a means of easily creating,
  training, and evaluating \nlp\ \dl\ models.
\end{abstract}

\zzsec{intro}{Introduction}

Consistently reproducing results is a fundamental criteria of the scientific
method, without which, a hypothesis may be weakened or even invalidated. This
is becoming even more necessary, because a growing number of publications are
inflated by false positives, to the point that they are labeled with the
pejorative term ``p-hacking'', the intentional or unintentional act to bias
results in favor of publication for peer
acceptance~\cite{headExtentConsequencesPHacking2015}; efforts in this direction
include introducing new statistical methods to detect false findings
~\cite{ulrichPhackingPostHoc2015}.

The inability to reproduce results has been referred to as the ``replication
crisis''~\cite{hutsonArtificialIntelligenceFaces2018}.  The problem of
reproducibility in results is becoming more acknowledged as a serious issue in
the \ml\ community with efforts to understand and overcome the
challenge~\cite{olorisadeReproducibilityMachineLearningBased2017}. Not only has
the community addressed the issue in the literature, it has endeavored to
assess if experiments are reproducible and provide recommendations to remedy
the problem where reproducibility is lacking.  An example of this effort
includes reporting on experimental methodology, implementation, analysis and
conclusions in the \reprodch.

To address these issues, we present \fname, a \dl\ framework for \nlp\ research
by and for the academic research community.  Not only does the framework
address issues of reproducibility, it also is designed to easily and quickly
test with varying model configurations such as extending contextual (and
non-contextual) word embeddings~\cite{devlinBERTPretrainingDeep2019,
  mikolovDistributedRepresentationsWords2013, penningtonGloveGlobalVectors2014}
with linguistic token level
features~\cite{huangBidirectionalLSTMCRFModels2015}, and join layer document
level
features~\cite{deerwesterIndexingLatentSemantic1990,sparckjonesStatisticalInterpretationTerm1972}
using easy to write configuration with little to no code.

What sets \fname\ apart from other frameworks is its capability of reproducing
results, efficient mini-batch creation for feature swapping for model
comparisons, and an emphasis on vectorization of natural language text
providing zero coding \nn\ construction.  The framework was written with
\nlp\ researchers, science related outcomers, and students in mind.

The framework's source code and installable libraries are released under the
\mitlic, which is available both on GitHub and as Python \texttt{pip} packages
along with extensive and in depth documentation, tutorials and \jupnoteintro\
examples.  The \api\ documentation is fully hyperlinked, includes overview
documentation, class diagrams, and tutorials.  The framework is validated with
236 unit tests and six integration tests, most of which are automated using
continuous integration testing for both functionality and reproducibility.

\zzsec{prev}{Previous Work}

Popular \dl\ frameworks such as \tenfl\ have a dashboard that provides metrics,
such as training and validation loss.  However, these general purpose
frameworks offer basic performance metrics and do not provide a means of
producing higher abstraction level \nlp\ specific models.  More specifically,
frameworks such as Keras, supply a very coarse \api\ allowing solely for
cookie-cutter models.  They lack the ability to easily create and evaluate
models past this surface interface.

Frameworks such as \pytintro, which are more common in academia, provide a more
straightforward simple \api\ that is similar to the core \tenfl\ libraries, and
thus have the same shortcomings as a tool to bridge the gap between pure
research and reproducibility.

\allenintro~\cite{gardnerAllenNLPDeepSemantic2018} is a flexible configuration
driven framework that provides ease of construction of \nlp\ \nn\
architectures, and thus, is the closest framework to ours.  However, it does
not have fast feature swapping \zzseesec{persist} and batch creation
capability, and lacks most of the components necessary to \allenres.

Popular packages providing support for transformer architectures such as
BERT~\cite{devlinBERTPretrainingDeep2019} include \hugface.  However, this
framework only provides transformer models for contextual word embeddings.

\zzsec{design}{Design}

Like the DeepDIVA~\cite{albertiDeepDIVAHighlyFunctionalPython2018}, \fname\ is
written in Python and utilizes, but does not replace, \pyt.  The goal of our
framework is:
\begin{zzpackeditem}
\item Reproducibility of results \zzseesec{reproduce},
\item Efficiently create and load mini-batches \zzseesec{persist},
\item Decouple the process of vectorizing data for reuse in \nn\ architectures
  \zzseesec{vec},
\item Provide language specific vectorization \zzseesec{nlfeat} and \dl\ layers
  \zzseesec{layers}.
\end{zzpackeditem}

\zzsubsec{reproduce}{Reproducibility}

All random state, including utility libraries, scientific libraries, the \pyt\
library, and GPU state, is consistent across each run of a Python interpreter
execution of the model's training, evaluation and testing.  Results consistency
is retained by saving this random state when saving the model, then
retrieving and resetting it after loading the model.

The order of mini-batches, and their constituent data can affect the model
performance as an aspect of training or the results of validation and testing.
This performance inconsistency is addressed by recording the order of all
data\footnote{Regardless of any user given data pre-processing or shuffling.}
and tracking the training, validation and test data splits.  Not only are
mini-batches given in the same order, the ordering in each mini-batch is also
preserved.  These dataset partitions and their order is saved to the file
system so the community has the option of distributing it along with the source
code for later experiment duplication.

In addition, the framework also saves the configuration used to recreate the
same in memory state along with the model.  This duplicates the model
structure, parameters, hyper-parameters and all other train-time memory during
testing.

\zzsubsec{stack}{Technology Stack}

Each ``layer'' of the stack builds on more general libraries to reduce the
installation footprint based on the needs of each use case.  Each library
contains the requirements for dependent third-party and lower tier packages.
The framework consists of the following libraries (see \zzfigref{\stackref}):
\zzfigure{\columnwidth}{\stackref}{The Python Library Stack}
\begin{itemize}
\item \zutil: Utility library command line parsing, persistence and a
  \javaspring\ like inversion of
  control~\cite{mattssonObjectOrientedFrameworks1996} configuration system.

\item \znlparse: Parses natural language text using \spacyintro, generates and
  vectorizes features.

\item \zdeeplearn: General purpose \dl\ \api, much like DeepDIVA, providing
  mini-batching, vectorization, and training, validation and testing of a
  model.

\item \zdeepnlp: Contains language vectorization, such as word embeddings
  \zzseesec{layers}, part of speech tags, named entity recognition, and head
  dependencies~\cite{mcdonaldOnlineLargemarginTraining2005}.
\end{itemize}

\zzsubsec{exec}{Execution}

The framework provides both a command line and a \jupytername\ interface to
train, test and predict.  A ``glue'' \api\ is used to make a \pdataclass\ class
a dynamically generated command line with help usage message documentation.  A
set of default application classes are available with the framework, but they
can be extended to include project specific actions.  The default application
set provides interactive early stopping or epoch resetting during training.

The command line and \jupytername\ both use a common facade interface to the
model itself, which is conducive as an entry point to both larger projects or
simple run scripts.  However, the \jupytername\ interface provides evaluation
training and validation loss plots \zzseefig{loss-snapshot}.

\zzfigure{\columnwidth}{loss-snapshot}{Validation and Training Loss
  Plot from the NER Token Classification Application}

Both interfaces provide a debugging mode that outputs a step of the model
training with batch composition, layer names, dimension calculations, using the
Python logging system, which is filterable by module or component.

\zzsubsec{vec}{Vectorization}

The \fname\ framework provides easily configurable components to digitally
vectorize data, which in our framework, is encapsulated in a
\textit{vectorizer}, which takes a particular input data and outputs a tensor.

The \zdeeplearnname\ library is general purpose with respect to any kind of
data, such as: text, images, audio.  Examples of data it can vectorize include
\pandas\ data frames, enumerated nominal values (typically used for labels),
and hot encoded vectors.

The \zdeepnlpname\ library provides a higher abstraction that parses natural
language text, sentence chunks, and vectorizes linguistic features.  These
vectorizers fall into one of the following categories:
\begin{zztypedefinition}{01}{\it}{0cm}
\item[token] Features taken from each token, shape congruent with the number of
  tokens, typically concatenated with the embedding layer include \spacy\
  features such as \pos\ tags, \ner\ tags, dependency tree tags and the depth
  of a token in its head dependency tree.
\item[document] Features taken from the document level, typically added to a
  join layer such as count sums of \spacy\ parsed features.
\item[multi-document] Aggregating and shared features between
  more than one document, such as overlapping \pos\ or \ner\ tags.
\item[embedding] Vectorizes text into word embeddings, such as sentence or
  document text.
\end{zztypedefinition}

See \zzsecref{nlfeat} for more information on \nlp\ specific feature generation.

\zzfigure{\columnwidth}{batch-encode-tag}{Parse Sentence Chunk and POS Tagged}

For example, suppose the following sentences are to be vectorized: {\it ``The boy
  hit the ball.  He did it well.''}.  First the \zparsename\ library is used to
parse, chunk and \pos\ tag the sentence (see \zzfigref{batch-encode-tag}), then
\pos\ tags are converted to one-hot encoded vectors by the \spacy\ feature
vectorizer (see \zzfigref{batch-encode-vectorize-2}).

\zzfigure{\columnwidth}{batch-encode-vectorize-2}{Vectorize Language}

\zzsubsec{persist}{Batching and Persistence}

Most {\nn}s expect and perform well using
mini-batches~\cite{ioffeBatchNormalizationAccelerating2015a} as input accept or
require batched input.  A na\"ive approach to generating these mini-batches
would be to re-parse and re-vectorize the data for each mini-batch over each
epoch, which is inefficient.  Many projects address this inefficiency by
pre-processing the data before training.  However, this leads to a brittle and
difficult to reproduce dataset generated set of ad-hoc text processing
scripts that are challenging re-execute, and thus, reproduce performance
metrics.

A cleaner and more efficient method is to wrap this process in the framework
and create a file system scheme with the intermediate files in a configured
location so as to not clutter the project workspace, which is supported by
\fname.

Another desirable feature of any framework is to easily swap in and out feature
sets and compare performance metrics, which usually takes the form of the
following steps:
\begin{zzpackedinline}
\item decide which features to use at train time,
\item train and evaluate,
\item test,
\item evaluate performance,
\item choose a different feature set,
\item go to step \textit{a}.
\end{zzpackedinline}
This incremental process highlights the need to efficiently create mini-batches.

A key observation is that each mini-batch is independent\footnote{There are
  exceptions for some algorithms that need to index and fit the corpus before
  vectorization.}.  The \zdeeplearnname\ leverages the fact that mini-batches
are independent and fit nicely as independent units of work by segmenting
datasets into smaller chunks, vectorizing each chunk in parallel sub
processes, and creating one or more batches independently across each sub
processes.  This process by which data is written to the file system in a
format that is fast to reassemble is called \textit{batch encoding}, and
accomplished by:
\begin{zzpackedenum}
\item Collecting data needed to vectorize,
\item Chunking data in to equal size parts,
\item Forking processes using the python multiprocessing api,
\item Vectorizing each chunk in each sub-process by:
  \begin{enumerate}
  \item Recreating parent memory by using configuration factory in the
    \zutilname\ library,
  \item Vectorize each data chunk as separate feature sets,
  \item Groups the vectorized in to bundles as mini-batches, but in separate files,
  \item Vectorized data is almost always ready-to-go tensors.
  \end{enumerate}
\end{zzpackedenum}

\textit{Batch decoding} is the process by which data is grouped for training as
mini-batches and is accomplished by:
\begin{zzpackedenum}
\item Choosing a feature set for a training run,
\item Reassembling features by mini-batch, then feature as a two level directory
  structure (see \zzfigref{batch-reassembly}),
\item Decode each mini-batch in to a tensor, if not unserialized from the file
  system as a tensor (see \zzfigref{batch-decode}),
\item Copy tensors to the GPU if available,
\item Cache tensors in CPU or GPU memory\footnote{Cached resources are tracked
    so GPU memory is maintained.}.
\end{zzpackedenum}

\zzfigure{\columnwidth}{batch-reassembly}{Batch Reassembly Process}

Reassembling mini-batches by feature greatly reduces load time and memory
space, which speeds up model training and allows for more complex models.  This
leverage is most apparent when comparing pre-generated frozen large BERT model
embeddings for frozen transformers compared to a trainable model.  In the case
of the former, large data files with output tensors are read back in compared
to word piece embeddings~\cite{wuGoogleNeuralMachine2016} for a trainable
model.

After mini-batch encoding is complete, several feature combinations can be created
in configuration, then trained, validated and tested offline.  Utility methods
exist to aggregate results in tabular form for reporting.

\zzfigure{\columnwidth}{batch-decode}{Batch Decode}

\zzsubsec{nlfeat}{Natural Language Features}

\zzfiguretc[5in]{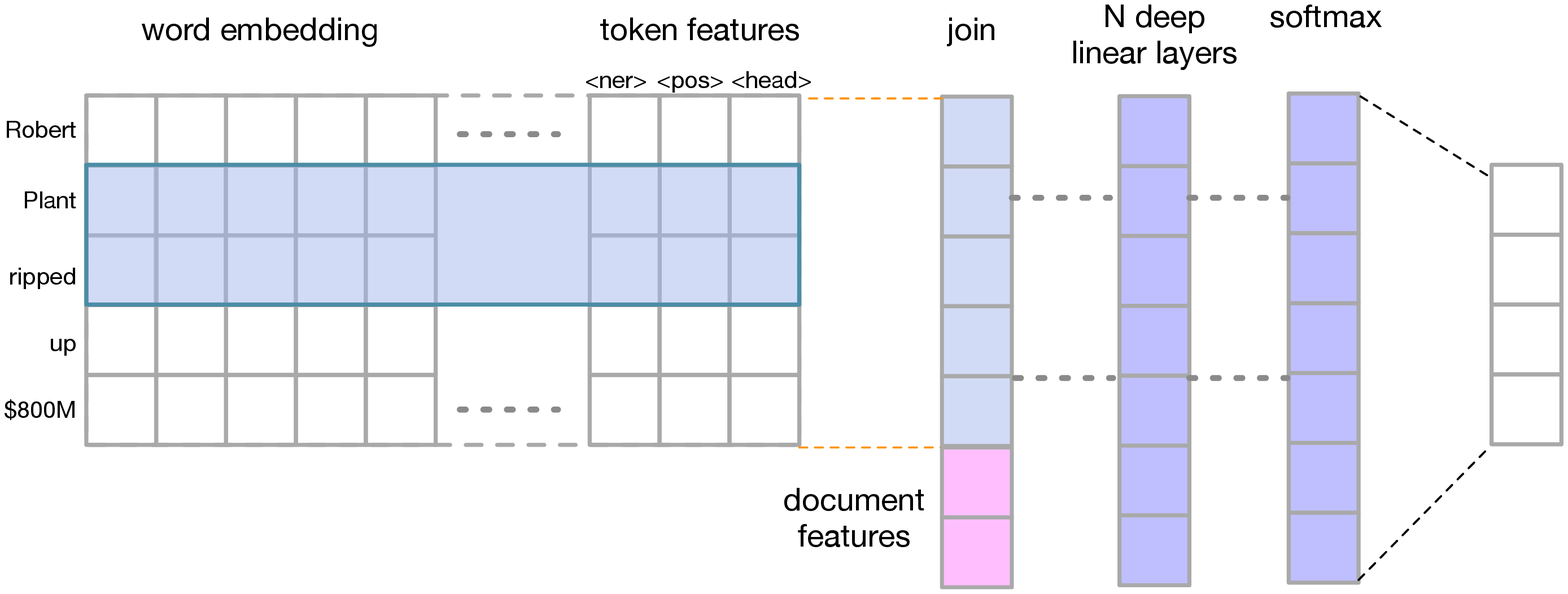}{Embedding Concatenation}

What sets \fname\ apart from other frameworks is not only efficient mini-batch
creation and feature swapping (\zzsecref{persist}), but the tight coupling of
natural language features with a deep learning \api. Specifically,
vectorization of natural language features is at the heart of the utility of
the \fname\ framework, and addresses a need that is otherwise lacking in other
\apis.

One such powerful capability is the concatenation of any vectorized data to
word embeddings, which is available for both contextual embeddings such as
BERT~\cite{devlinBERTPretrainingDeep2019} and non-contextual embeddings such as
GLoVE~\cite{penningtonGloveGlobalVectors2014} \zzseesec{layers} for supported
word embedding types.

In addition to concatenation of word embeddings, document level features can be
added to a join layer (see \zzfigref{deepnlp-embedding}).

\zzsubsec{layers}{Layers}

The framework provides many layer implementations, which extend from the \pyt\
\texttt{Module} class, thus any \pyt\ module can be used in the framework.  To
this end, it uses layers such as the \pytorchcrf\ conditional random field
implementation to create an end-to-end model for sequence classification.
Layers are configured in memory to offload the construction details to the
framework.  Other layers provided include, but are not limited to:
\begin{zzpackeditem}
\item BiLSTM CRF for applications such as sequence
  tagging~\cite{huangBidirectionalLSTMCRFModels2015} as an end-to-end model,
  which requires no coding with the exception of mapping data input to
  vectorizers.
\item BERT transformer models for sentence and token classification.
\item 1D convolution \nn\ that provides calculation for an arbitrarily deep
  network with input and output dimensionality calculation, pooling,
  mini-batch~\cite{ioffeBatchNormalizationAccelerating2015a} centering and
  activation.
\item Expanding or contracting \dl\ feed forward linear networks with repeats
  with input and output feature calculation.
\item Word embedding layer capable of concatenating linguistic features at both
  the token and document level \zzseesec{vec}.  Supported embeddings include
  Bert~\cite{devlinBERTPretrainingDeep2019},
  word2vec~\cite{mikolovDistributedRepresentationsWords2013},
  GLoVE~\cite{penningtonGloveGlobalVectors2014} and
  FastText~\cite{bojanowskiEnrichingWordVectors2017} \zzseesec{nlfeat}.
\item Document level \lsa~\cite{deerwesterIndexingLatentSemantic1990}, and term
  frequency with inverse document frequency
  weighting~\cite{sparckjonesStatisticalInterpretationTerm1972} are also
  available out of the box with no coding required
\end{zzpackeditem}

HuggingFace transformer features are available as an embeddings vectorizer and
document, sentence and token classification are available as layers.  In
addition a vectorizer is provided to map linguistic features created by
\spacyname\ to word piece embeddings~\cite{wuGoogleNeuralMachine2016} for
concatenation with the last hidden state of the transformer.

\zzsec{limits}{Limitations and Future Work}

While the framework is thoroughly tested in the areas it was designed for, some
work remains to enable more variations in \dl\ architecture for \sota\
experimentation.

Any of the \hugfacemodels\ are available, but only
BERT~\cite{devlinBERTPretrainingDeep2019},
RoBERTa~\cite{liuRoBERTaRobustlyOptimized2019} and
DistilBERT~\cite{sanhDistilBERTDistilledVersion2020} have been tested and next
sentence and masking prediction is not yet implemented.

A planned future work is to integrate the framework with TensorFlow's \tb,
which provides real-time graphing of metrics such as training and validation
loss.

\zzsec{conc}{Conclusion}

The \fname\ framework has been presented as a viable solution to easily create
\nlp\ specific models with \apis\ and analysis tools to produce consistent
results.  Such frameworks are not only necessary, but vital in order to ensure
the legitimacy of the area of \dl\ in \nlp\ by providing the means necessary
to produce reliable reproducible results.

\bibliographystyle{acl_natbib}
\interlinepenalty=10000
\bibliography{DeepZensols.bib}

\begin{thebibliography}{19}
\expandafter\ifx\csname natexlab\endcsname\relax\def\natexlab#1{#1}\fi

\bibitem[{Alberti et~al.(2018)Alberti, Pondenkandath, Würsch, Ingold, and
  Liwicki}]{albertiDeepDIVAHighlyFunctionalPython2018}
Michele Alberti, Vinaychandran Pondenkandath, Marcel Würsch, Rolf Ingold, and
  Marcus Liwicki. 2018.
\newblock \href {https://doi.org/10.1109/ICFHR-2018.2018.00080} {{{DeepDIVA}}:
  {{A Highly}}-{{Functional Python Framework}} for {{Reproducible
  Experiments}}}.
\newblock In \emph{2018 16th {{International Conference}} on {{Frontiers}} in
  {{Handwriting Recognition}} ({{ICFHR}})}, pages 423--428.

\bibitem[{Bojanowski et~al.(2017)Bojanowski, Grave, Joulin, and
  Mikolov}]{bojanowskiEnrichingWordVectors2017}
Piotr Bojanowski, Edouard Grave, Armand Joulin, and Tomas Mikolov. 2017.
\newblock \href {https://transacl.org/ojs/index.php/tacl/article/view/999}
  {Enriching {{Word Vectors}} with {{Subword Information}}}.
\newblock \emph{Transactions of the Association for Computational Linguistics},
  5:135--146.

\bibitem[{Deerwester et~al.(1990)Deerwester, Dumais, Furnas, Landauer, and
  Harshman}]{deerwesterIndexingLatentSemantic1990}
Scott Deerwester, Susan~T. Dumais, George~W. Furnas, Thomas~K. Landauer, and
  Richard Harshman. 1990.
\newblock Indexing by {{Latent Semantic Analysis}}.
\newblock \emph{Journal of the American Society for Information Science; New
  York, N.Y.}, 41(6):391--407.

\bibitem[{Devlin et~al.(2019)Devlin, Chang, Lee, and
  Toutanova}]{devlinBERTPretrainingDeep2019}
Jacob Devlin, Ming-Wei Chang, Kenton Lee, and Kristina Toutanova. 2019.
\newblock \href {https://doi.org/10.18653/v1/N19-1423} {{{BERT}}:
  {{Pre}}-training of {{Deep Bidirectional Transformers}} for {{Language
  Understanding}}}.
\newblock In \emph{Proceedings of the 2019 {{Conference}} of the {{North
  American Chapter}} of the {{Association}} for {{Computational Linguistics}}:
  {{Human Language Technologies}}, {{Volume}} 1 ({{Long}} and {{Short
  Papers}})}, pages 4171--4186.

\bibitem[{Gardner et~al.(2018)Gardner, Grus, Neumann, Tafjord, Dasigi, Liu,
  Peters, Schmitz, and Zettlemoyer}]{gardnerAllenNLPDeepSemantic2018}
Matt Gardner, Joel Grus, Mark Neumann, Oyvind Tafjord, Pradeep Dasigi,
  Nelson~F. Liu, Matthew~E. Peters, Michael Schmitz, and Luke Zettlemoyer.
  2018.
\newblock \href {https://doi.org/10.18653/v1/W18-2501} {{{AllenNLP}}: {{A Deep
  Semantic Natural Language Processing Platform}}}.
\newblock In \emph{Proceedings of {{Workshop}} for {{NLP Open Source Software}}
  ({{NLP}}-{{OSS}})}, pages 1--6.

\bibitem[{Head et~al.(2015)Head, Holman, Lanfear, Kahn, and
  Jennions}]{headExtentConsequencesPHacking2015}
Megan~L. Head, Luke Holman, Rob Lanfear, Andrew~T. Kahn, and Michael~D.
  Jennions. 2015.
\newblock \href {https://doi.org/10.1371/journal.pbio.1002106} {The {{Extent}}
  and {{Consequences}} of {{P}}-{{Hacking}} in {{Science}}}.
\newblock \emph{PLoS Biology}, 13(3).

\bibitem[{Huang et~al.(2015)Huang, Xu, and
  Yu}]{huangBidirectionalLSTMCRFModels2015}
Zhiheng Huang, Wei Xu, and Kai Yu. 2015.
\newblock \href {http://arxiv.org/abs/1508.01991} {Bidirectional {{LSTM}}-{{CRF
  Models}} for {{Sequence Tagging}}}.
\newblock arXiv: 1508.01991.

\bibitem[{Hutson(2018)}]{hutsonArtificialIntelligenceFaces2018}
Matthew Hutson. 2018.
\newblock \href {https://doi.org/10.1126/science.359.6377.725} {Artificial
  intelligence faces reproducibility crisis}.
\newblock \emph{Science}, 359(6377):725--726.

\bibitem[{Ioffe and Szegedy(2015)}]{ioffeBatchNormalizationAccelerating2015a}
Sergey Ioffe and Christian Szegedy. 2015.
\newblock \href {http://arxiv.org/abs/1502.03167} {Batch {{Normalization}}:
  {{Accelerating Deep Network Training}} by {{Reducing Internal Covariate
  Shift}}}.
\newblock In \emph{Proceedings of the 32nd {{International Conference}} on
  {{International Conference}} on {{Machine Learning}} - {{Volume}} 37},
  {{ICML}}'15, pages 448--456. {JMLR.org}.

\bibitem[{Liu et~al.(2019)Liu, Ott, Goyal, Du, Joshi, Chen, Levy, Lewis,
  Zettlemoyer, and Stoyanov}]{liuRoBERTaRobustlyOptimized2019}
Yinhan Liu, Myle Ott, Naman Goyal, Jingfei Du, Mandar Joshi, Danqi Chen, Omer
  Levy, Mike Lewis, Luke Zettlemoyer, and Veselin Stoyanov. 2019.
\newblock \href {http://arxiv.org/abs/1907.11692} {{{RoBERTa}}: {{A Robustly
  Optimized BERT Pretraining Approach}}}.
\newblock arXiv: 1907.11692.

\bibitem[{Mattsson(1996)}]{mattssonObjectOrientedFrameworks1996}
Michael Mattsson. 1996.
\newblock \href
  {http://citeseerx.ist.psu.edu/viewdoc/download?doi=10.1.1.36.1424&rep=rep1&type=pdf}
  {Object-{{Oriented Frameworks}}}.
\newblock masters, {Lund University}, {Ronneby, Sweden}.

\bibitem[{McDonald et~al.(2005)McDonald, Crammer, and
  Pereira}]{mcdonaldOnlineLargemarginTraining2005}
Ryan McDonald, Koby Crammer, and Fernando Pereira. 2005.
\newblock \href {https://doi.org/10.3115/1219840.1219852} {Online
  {{Large}}-margin {{Training}} of {{Dependency Parsers}}}.
\newblock In \emph{Proceedings of the 43rd {{Annual Meeting}} on
  {{Association}} for {{Computational Linguistics}}}, {{ACL}} '05, pages
  91--98. {Association for Computational Linguistics}.

\bibitem[{Mikolov et~al.(2013)Mikolov, Sutskever, Chen, Corrado, and
  Dean}]{mikolovDistributedRepresentationsWords2013}
Tomas Mikolov, Ilya Sutskever, Kai Chen, Greg~S Corrado, and Jeff Dean. 2013.
\newblock \href
  {http://papers.nips.cc/paper/5021-distributed-representations-of-words-and-phrases-and-their-compositionality.pdf}
  {Distributed {{Representations}} of {{Words}} and {{Phrases}} and their
  {{Compositionality}}}.
\newblock In C.~J.~C. Burges, L.~Bottou, M.~Welling, Z.~Ghahramani, and K.~Q.
  Weinberger, editors, \emph{Advances in {{Neural Information Processing
  Systems}} 26}, pages 3111--3119. {Curran Associates, Inc.}

\bibitem[{Olorisade et~al.(2017)Olorisade, Brereton, and
  Andras}]{olorisadeReproducibilityMachineLearningBased2017}
Babatunde~K. Olorisade, Pearl Brereton, and Peter Andras. 2017.
\newblock \href {https://openreview.net/forum?id=By4l2PbQ-} {Reproducibility in
  {{Machine Learning}}-{{Based Studies}}: {{An Example}} of {{Text Mining}}}.

\bibitem[{Pennington et~al.(2014)Pennington, Socher, and
  Manning}]{penningtonGloveGlobalVectors2014}
Jeffrey Pennington, Richard Socher, and Christopher Manning. 2014.
\newblock \href {http://www.aclweb.org/anthology/D14-1162} {Glove: {{Global
  Vectors}} for {{Word Representation}}}.
\newblock In \emph{Proceedings of the 2014 {{Conference}} on {{Empirical
  Methods}} in {{Natural Language Processing}} ({{EMNLP}})}, pages 1532--1543.
  {Association for Computational Linguistics}.

\bibitem[{Sanh et~al.(2020)Sanh, Debut, Chaumond, and
  Wolf}]{sanhDistilBERTDistilledVersion2020}
Victor Sanh, Lysandre Debut, Julien Chaumond, and Thomas Wolf. 2020.
\newblock \href {http://arxiv.org/abs/1910.01108} {{{DistilBERT}}, a distilled
  version of {{BERT}}: Smaller, faster, cheaper and lighter}.
\newblock arXiv: 1910.01108.

\bibitem[{Sparck~Jones(1972)}]{sparckjonesStatisticalInterpretationTerm1972}
Karen Sparck~Jones. 1972.
\newblock A statistical interpretation of term specificity and its application
  in retrieval.
\newblock \emph{Journal of documentation}, 28(1):11--21.

\bibitem[{Ulrich and Miller(2015)}]{ulrichPhackingPostHoc2015}
Rolf Ulrich and Jeff Miller. 2015.
\newblock \href
  {https://doi.org/http://dx.doi.org.proxy.cc.uic.edu/10.1037/xge0000086}
  {P-hacking by post hoc selection with multiple opportunities:
  {{Detectability}} by skewness test?: {{Comment}} on {{Simonsohn}},
  {{Nelson}}, and {{Simmons}} (2014)}.
\newblock \emph{Journal of Experimental Psychology: General},
  144(6):1137--1145.

\bibitem[{Wu et~al.(2016)Wu, Schuster, Chen, Le, Norouzi, Macherey, Krikun,
  Cao, Gao, Macherey, Klingner, Shah, Johnson, Liu, Kaiser, Gouws, Kato, Kudo,
  Kazawa, Stevens, Kurian, Patil, Wang, Young, Smith, Riesa, Rudnick, Vinyals,
  Corrado, Hughes, and Dean}]{wuGoogleNeuralMachine2016}
Yonghui Wu, Mike Schuster, Zhifeng Chen, Quoc~V. Le, Mohammad Norouzi, Wolfgang
  Macherey, Maxim Krikun, Yuan Cao, Qin Gao, Klaus Macherey, Jeff Klingner,
  Apurva Shah, Melvin Johnson, Xiaobing Liu, Łukasz Kaiser, Stephan Gouws,
  Yoshikiyo Kato, Taku Kudo, Hideto Kazawa, Keith Stevens, George Kurian,
  Nishant Patil, Wei Wang, Cliff Young, Jason Smith, Jason Riesa, Alex Rudnick,
  Oriol Vinyals, Greg Corrado, Macduff Hughes, and Jeffrey Dean. 2016.
\newblock \href {http://arxiv.org/abs/1609.08144} {Google's {{Neural Machine
  Translation System}}: {{Bridging}} the {{Gap}} between {{Human}} and
  {{Machine Translation}}}.
\newblock arXiv: 1609.08144.

\end{thebibliography}

\end{document}